\documentclass[10pt, conference]{IEEEtran}

\usepackage{amsmath,amssymb,amsfonts}
\usepackage{graphicx}
\usepackage{pifont}

\usepackage{amsthm}
\usepackage{xcolor}
\usepackage[figuresright]{rotating}

\usepackage{graphicx}
\graphicspath{{/}{fig/}}

\usepackage{array}
\usepackage{textcomp}
\usepackage{xcolor}
\usepackage{multirow}
\usepackage{booktabs}

\usepackage{mathtools}
\usepackage{breqn}
\usepackage{float}

\usepackage{pgfplots}
\pgfplotsset{compat=1.7}
\usepgfplotslibrary{groupplots}

\usepackage{caption}
\usepackage{subcaption}

\usepackage{multirow,tabularx}
\usepackage{hyperref}
\usepackage{flushend}
\usepackage{algorithmic}
\usepackage[vlined, ruled, shortend]{algorithm2e}

\newlength\figureheight
\newlength\figurewidth
\SetAlCapNameFnt{\footnotesize}
\SetAlCapFnt{\footnotesize}

\DeclareMathOperator*{\argmin}{argmin}

\title{
    Cooperative UWB-Based Localization for Outdoors Positioning and Navigation of UAVs aided by Ground Robots \\
}

\author{
    \IEEEauthorblockN{
        \vspace{1em}
        Yu Xianjia\IEEEauthorrefmark{2},
        Li Qingqing\IEEEauthorrefmark{2},
        Jorge Peña Queralta\IEEEauthorrefmark{2},
        Jukka Heikkonen\IEEEauthorrefmark{2},
        Tomi Westerlund\IEEEauthorrefmark{2}
    }
    \IEEEauthorblockA{
        \normalsize
        \IEEEauthorrefmark{2}\href{https://tiers.utu.fi}{Turku Intelligent Embedded and Robotic Systems (TIERS) Lab, University of Turku, Finland}.\\
        Emails: \textsuperscript{1}\{xianjia.yu, qingqli, jopequ, jukhei, tovewe\}@utu.fi\\[+6pt]
    }
}

\begin{document}

\maketitle
\thispagestyle{empty}
\pagestyle{empty}



\begin{abstract}%
    \label{sec:abstract}%
    Unmanned aerial vehicles (UAVs) are becoming largely ubiquitous with an increasing demand for aerial data. Accurate navigation and localization, required for precise data collection in many industrial applications, often relies on RTK GNSS. These systems, able of centimeter-level accuracy, require a setup and calibration process and are relatively expensive. This paper addresses the problem of accurate positioning and navigation of UAVs through cooperative localization. Inexpensive ultra-wideband (UWB) transceivers installed on both the UAV and a support ground robot enable centimeter-level relative positioning. With fast deployment and wide setup flexibility, the proposed system is able to accommodate different environments and can also be utilized in GNSS-denied environments. Through extensive simulations and test fields, we evaluate the accuracy of the system and compare it to GNSS in urban environments where multipath transmission degrades accuracy. For completeness, we include visual-inertial odometry in the experiments and compare the performance with the UWB-based cooperative localization.

\end{abstract}

\IEEEpeerreviewmaketitle


\section{Introduction}\label{sec:introduction}

Multiple industrial use cases benefit from the deployment of Unmanned aerial vehicles (UAVs)~\cite{shakhatreh2019unmanned}. When accurate localization is needed, GNSS-RTK is the de-facto standard for gathering aerial data with UAVs~\cite{li2018high}. For example, high-accuracy photogrammetry~\cite{lee2018assessment}, civil infrastructure monitoring~\cite{kim2018structural}, or in urban environments where GNSS signals suffer more degradation~\cite{li2018high}. As UAVs become ubiquitous across different domains and application areas~\cite{queralta2020collaborative}, having access to more flexible and lower-cost solutions to precise UAV navigation can aid in accelerating adoption and widespread use. In this paper, we consider the problem of UAV navigation through relative localization to a companion unmanned ground vehicle (UGV). We consider a ground robot as a more flexible platform from the point of view of deployment, but in simulations, we also consider localization based on fixed beacons in the environment, closer to how GNSS-RTK systems are deployed.

Within the different approaches that can be used for cooperative relative localization, from visual sensors~\cite{hui2013autonomous} to cooperative SLAM~\cite{kim2019uav}, wireless ranging technologies offer high performance with low system complexity~\cite{queralta2020uwb}. In particular, ultra-wideband (UWB) wireless ranging offers unparalleled localization performance within the different radio technologies in unlicensed bands~\cite{shule2020uwb}. Other benefits of UWB include resilience to multipath, high time resolution, and low interference with other radio technologies~\cite{yu2021applications}.

\begin{figure}
    \centering
    \includegraphics[width=0.49\textwidth]{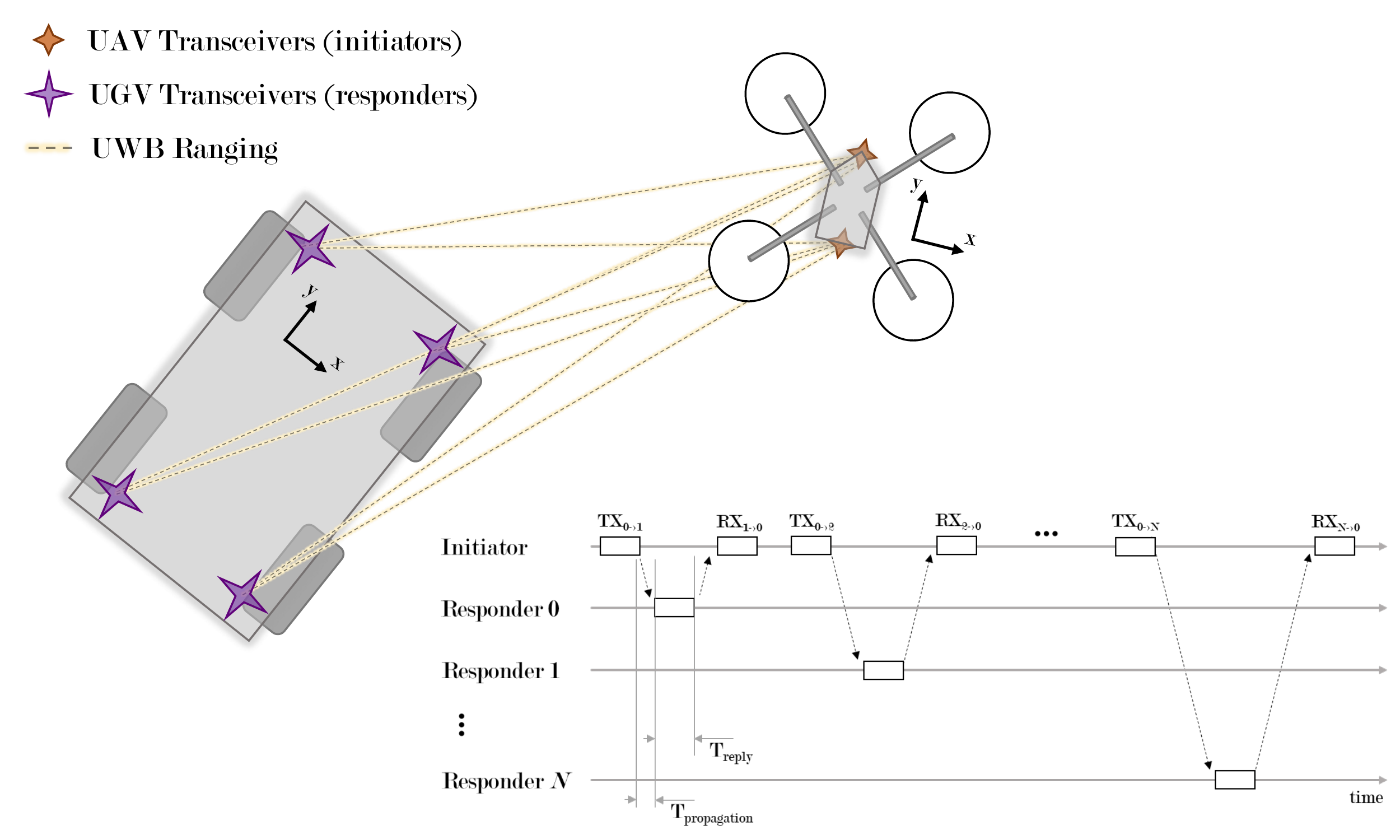}
    \caption{Cooperative localization approach based on UWB ranging measurements from multiple transceivers in different robots}
    \label{fig:concept}
\end{figure}

The system we analyze in this paper consists of a UGV equipped with four UWB transceivers and a UAV equipped with two transceivers. The UAV transceivers act as initiators, taking turns in sending signals to each of the UGV transceivers. When these respond, the time of flight of the signal is calculated and the distance between each pair of transceivers is calculated. This process is illustrated in Fig.~\ref{fig:concept}. The main contribution of this paper is thus on evaluating how UWB-based relative localization can improve the positioning of UAVs when supported by ground robots. We simulate different trajectories to evaluate the performance of the system and compare the accuracy of the GNSS, UWB, and VIO approach to localization with field tests in an urban environment. In the simulations, we consider different configurations of transceivers in the ground to compare the localization and navigation performance.

The remainder of this document is organized as follows. Section II introduces absolute and relative positioning approaches relevant to the presented approach. Section III then describes the cooperative localization approach. In Section IV we introduce the methodology for simulations and experiments, with results presented in Section V. Section VI concludes the work and outlines future research directions.


\section{Background} \label{sec:related_work}

This section reviews the literature in the area of outdoors positioning and navigation methods for multi-robot systems. 

\subsection{Limitations of standalone GNSS}

The long-term operation of autonomous robots outdoors is often a reliance on GNSS~\cite{qingqing2019multi}.  However, the positioning accuracy of GNSS can be easily influenced by multipath when satellites are not in line-of-sight. This is a typical problem in urban environments or partly covered environments such as forests~\cite{groves2012intelligent, li2020localization}. Additional sensors are thus used in practice, from IMUs at the lowest level~\cite{jiang2017board} to odometry estimation from lidars~\cite{chang2019gnss} or visual sensors~\cite{li2019tight}. It is worth mentioning, nonetheless, that more recent receivers exploiting multi-constellation signals (e.g., GPS, GLONASS, BEIDOU, or GALILEO) are able to deliver significantly higher positioning accuracy~\cite{li2019triple}.


\subsection{GNSS-RTK}
High-accuracy GNSS positioning is possible with real-time kinematic (RTK) systems. RTK positioning leverages measurements of the phase of the signal's carrier wave in addition to the information content of the signal and relies on a reference station or interpolated virtual station to provide real-time corrections, providing up to centimeter-level accuracy~\cite{tomavstik2019uav}. These systems, however, are costly and require calibration and setup for each different location.

\subsection{Onboard navigation}

With the increasing adoption of UAVs in recent years, the maturity of onboard estate estimation and localization has reached a point where it is standard in commercial systems. Onboard odometry and positioning are typically based on monocular or stereo vision (e.g., VINS-mono~\cite{qin2018vins}, vins-fusion~\cite{qin2019general}), but lidars are also effective in larger UAVs~\cite{zhang2019maximum}. Passive visual sensors, however, have evident limitations in terms of environmental conditions (e.g., night operation) and in situations where there is a lack of features~\cite{xiao2017uav, qingqing2020towards}

\subsection{UWB Localization}

Ultra-wideband (UWB) positioning systems are being increasingly adopted for autonomous systems ~\cite{yu2021applications}.UWB positioning systems based on a series of fixed nodes in known locations (or anchors), and ranging measurements between these and mobile nodes (or tags), can be used for consistent, long-term localization of mobile robots~\cite{macoir2019uwb, queralta2020uwb}.
Compared to RTK-based localization systems, UWB systems can be utilized both indoors and outdoors, can be automatically calibrated~\cite{almansa2020autocalibration} for ad-hoc deployment, and offer similar accuracies at much lower prices. UWB sensors also have a small form factor and are generally considered more energy efficient than other wireless solutions. Finally, UWB ranging is often combined with other sensors to add orientation estimation and increase the overall localization performance. Different approaches in the literature include fusion of UWB with IMU~\cite{yao2017integrated}, VIO estimators~\cite{nguyen2019integrated}, GPS~\cite{zhang2019combined}, or lidar~\cite{song2019uwb}.

 

\subsection{UWB for relative estimation}

UWB ranging has been widely used for relative localization within multi-robot systems. For instance, in~\cite{nguyen2018robust}, the authors demonstrate a system where relative positioning between and UAV and a UGV is designed based on UWB transceivers installed on both robots. In subsequent works~\cite{nguyen2019integrated}, a similar system is employed during docking maneuvers. Combined with vision sensors for the final docking, the autonomous approach of the UAV to the UGV relied on UWB ranging between transceivers in both robots. Relative localization between UAVs and UGVs has also been shown within the context of collaborative dense scene reconstruction~\cite{queralta2020vio}. In multi-UAV systems and UAV swarms, UWB ranging has been leveraged for swarm-level decentralized estate estimation~\cite{xu2020decentralized, qi2020cooperative}.

In general, terms, while UWB systems including those for relative localization have been widely studied in the literature, we see a lack of studies that quantitatively analyze how UWB-based relative estate estimation can improve GNSS positioning and navigation outdoors.






\section{Cooperative UWB-Based Localization}


We consider the problem of relative localization between a UAV and a UGV based on UWB ranging between transceivers installed onboard both robots. The objective is to leverage this relative localization to improve the accuracy of the UAV navigation outdoors. We are especially interested in improving the navigation performance in urban areas where the accuracy of GNSS sensors is degraded due to the signal being reflected at or occluded by nearby buildings.

Let us denote by $I = \{I_i\}_{i=0,\dots,N-1}$ the set of $N$ transceivers onboard the UAV. These will act as initiators, i.e., will actively transmit messages to initiate ranging measurements between them and the responder transceivers on the ground. We denote the latter ones by the set $R = \{R_0\}_{i=0,\dots,M-1}$. An initial approach, which we implement, is to iteratively range between each initiator and the set of responders. If the number of nodes increases significantly, more scalable approaches can be used where, for example, a single initiator message is answered by several or all responders with different delays~\cite{grobetawindhager2019snaploc}.

We model the UWB ranges between an initiator $i$ and a responder $j$ with
\begin{equation}
    \textbf{z}^{UWB}_{(i,j)} = \lVert \textbf{p}_i(t)-\textbf{q}_j(t) \lVert\:+\: \mathcal{N}\left(0,\:\sigma_{UWB}\right)
\end{equation}
where $\textbf{p}_i$ and $\textbf{q}_j$ represent the positions of the initiator and responder transceivers, respectively. Based on the ranges, different approaches to localization include, e.g., multilateration or a least squares estimator (LE). We implement the latter, and hence the position of each tag can be calculated based on the known anchor positions by
\begin{equation}
    \textbf{p}_{i} = \argmin_{\textbf{p}\in\mathbb{R}^3} \displaystyle\sum_{j=0}^{M} \left(\textbf{z}^{UWB}_{(i,j)} - \lVert \textbf{p} - \textbf{q}_j \lVert \right)^2 
\end{equation}
Alternatively, assuming that the position of initiators in the UAV ($\{\textbf{p}_i\}$) is given based on the UAV's position and orientation ($\textbf{p}$ and $\theta$, respectively) by a set of rigid body transformations $f_i$, i.e., $\textbf{p}_i = f_i\left(\textbf{p}, \theta\right)$, then the estimator can be used to obtain the full pose of the UAV directly with
\begin{equation}
    \textbf{p}, \:\theta = \argmin_{\substack{\textbf{p}\in\mathbb{R}^3\\\:\theta\in(-\pi,\pi]}} \displaystyle\sum_{i=0}^{N}\displaystyle\sum_{j=0}^{M} \left(\textbf{z}^{UWB}_{(i,j)} - \lVert f_i\left(\textbf{p}, \theta\right) - \textbf{q}_j \lVert \right)^2
\end{equation}

\section{Methodology}

This section describes the simulation settings and robotic platforms utilized in the field experiments.

\subsection{Simulation environment}

The first tests are carried out in a simulation environment using ROS and Gazebo. We simulate the UWB ranging with a standard deviation of the Gaussian noise set to $\sigma_{UWB} = 10\,cm$. This is a conservative value based on the literature~\cite{queralta2020uwb}. We simulate a single transceiver on the UAV and four transceivers on the ground. The latter ones are set are variable distances simulating deployment in small UGVs (0.6\, separation), large UGVs (1.2\,m separation) and different settings based, e.g., on tripods (with separations at 3\,m, 4\,m, 12\,m and 16\,m).

In the simulation experiments, we perform two types of flights. First, a vertical flight where the UAV is set to follow a straight vertical line up to an altitude of 30\,m. Second, a flight following a square pattern with a fixed size of 8 by 8\,m but at different altitudes (5\,m, 10\,m and 20\,m). For each of these flights, we evaluate the UWB positioning performance with flights based on ground truth positioning. Then, we perform the flight using the UWB position estimation as control input and evaluate how well the UAV follows the predefined trajectory (we refer to this as navigation error).



\begin{figure}
    \centering
    \includegraphics[width=0.49\textwidth]{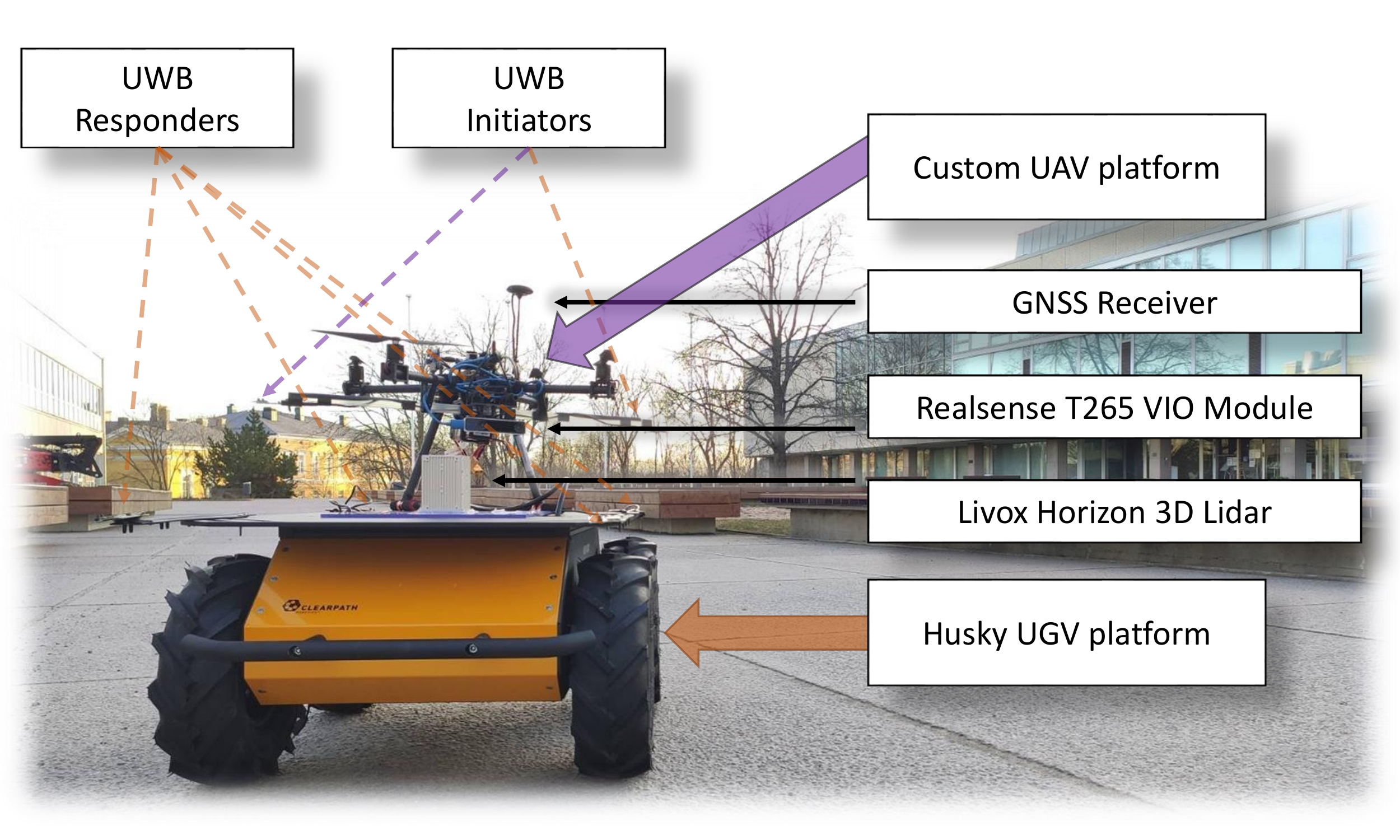}
    \caption{UAV and companion ground robot utilized in the experiments.}
    \label{fig:robots}
\end{figure}

\subsection{Multi-robot system}

The multi-robot system employed consists of a single ground robot and a UAV. The ground robot is a ClearPath Husky outdoor platform equipped with four UWB responder transceivers for cooperative positioning and a Livox Avia lidar utilized to obtain ground truth. Owing to the lack of a reference system such as a GNSS-RTK receiver, we extract the UAV position from the lidar's point cloud and utilize this as a reference. The point cloud is automatically processed following the steps described in 
Algorithm~\ref{alg:tracking}, and manually validated. We refer the reader to
~\cite{qingqing2021adaptive}
for further details on this method. Based on indoor testing with a reference anchor-based UWB system, we have evaluated the ground truth accuracy to be in the order of 10\,cm. The UGV and the custom UAV are shown in Fig.~\ref{fig:robots}. The UAV is equipped with two UWB transceivers and an Intel RealSense tracking camera T265 that performs VIO estimation.

\subsection{Experimental settings}

The field experiments are carried out in Turku, Finland (precise location is 60.4557389\textdegree\,N, 22.2843384\textdegree\,E), between a short line of trees and a large building that presumably blocks and reflects GNSS signals. The UAV runs the PX4 autopilot firmware, which is unable to obtain a stable GNSS lock in the test location. This location is chosen as an example of an urban location where GNSS receivers operate in suboptimal mode.

\begin{algorithm}[t]
    \footnotesize
	\caption{\footnotesize Ground truth extraction}
	\label{alg:tracking}
	\KwIn{\\
	    \begin{tabular}{ll}
	        3D lidar point cloud:             & $\mathcal{P}$ \\
	        Last known MAV state:              & $ \textbf{p}^{k-1}_{MAV}, {\dot{\textbf{p}}_{MAV}}^{k-1}$ \\[+0.4em]
	    \end{tabular}
	}
	\KwOut{\\
	    \begin{tabular}{ll}
	        MAV state:      & \{${\textbf{p}^{k}_{MAV}}, \:\dot{\textbf{p}}^{k}_{MAV}\}$ \\[+0.2em] 
	    \end{tabular}
	}  
	%
	\SetKwFunction{FSub}{$object\_extraction\left(\mathcal{P}, \:{\textbf{p}^{k-1}_{MAV}}, \:\dot{\textbf{p}}^{k-1}_{MAV}\right)$}
    \SetKwProg{Fn}{Function}{:}{}
    \BlankLine
            
    %
    \While{new $\mathcal{P}_{k}$}{
        \begin{tabular}{ll}
                Generate KD Tree:       & $kdtree \leftarrow \mathcal{P};$ \\
                MAV pos estimation:     & $\hat{\textbf{p}}_{\tiny MAV}^{k} \leftarrow \textbf{p}^{k-1}_{MAV} + \frac{\dot{\textbf{p}}_{MAV}^{k-1}}{I};$ \\[+0.4em]
                MAV points:             & $\mathcal{P}^{k}_{MAV} = KNN(kdtree, \: \hat{\textbf{p}}_{MAV}^{k});$ \\[+0.4em]
                MAV state estimation:   & ${\textbf{p}^{k}_{MAV}} = \frac{1}{\lvert\mathcal{P}^{k}_{MAV}\rvert}\sum_{p\in\mathcal{P}^{k}_{MAV}} p;$ \\
        \end{tabular}
    }

\end{algorithm}


\begin{figure*}
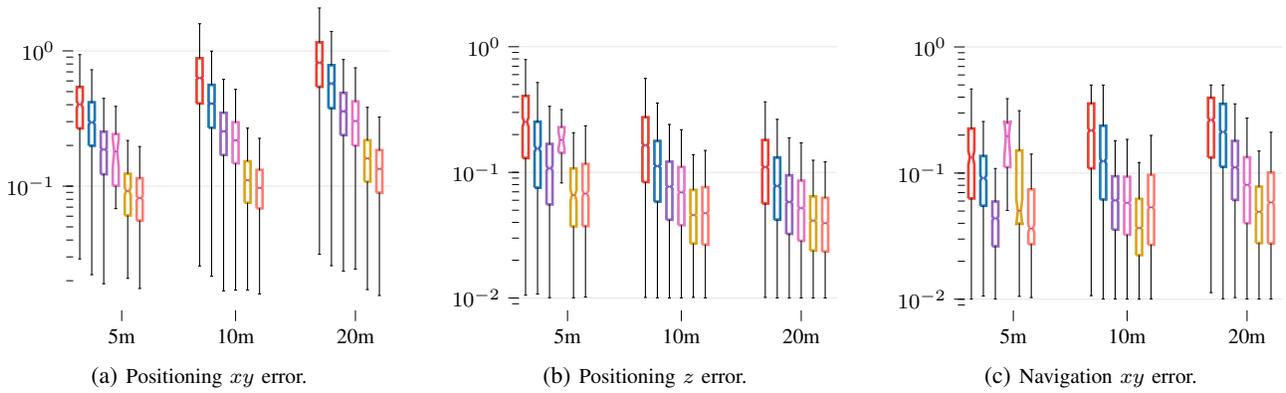

    \centering
    \begin{subfigure}{0.32\textwidth}
        \centering
        \setlength\figureheight{\textwidth}
        \setlength\figurewidth{\textwidth}
        \footnotesize{\input{tex/square_boxplot_position_xy}}
        \caption{Positioning $xy$ error.}
        \label{fig:square_pos_xy}
    \end{subfigure}
    \begin{subfigure}{0.32\textwidth}
        \centering
        \setlength\figureheight{\textwidth}
        \setlength\figurewidth{\textwidth}
        \footnotesize{\input{tex/square_boxplot_position_z}}
        \caption{Positioning $z$ error.}
        \label{fig:square_pos_z}
    \end{subfigure}
        \begin{subfigure}{0.32\textwidth}
        \centering
        \setlength\figureheight{\textwidth}
        \setlength\figurewidth{\textwidth}
        \footnotesize{\input{tex/square_boxplot_navi_xy}}
        \caption{Navigation $xy$ error.}
        \label{fig:square_nav}
    \end{subfigure}
    \caption{Positioning and navigation errors over a flight following a squared shape of 8 by 8\,m, at three different altitudes (5, 10 and 20\,m). The altitude is set to a constant so only the XY error is calculated for the UWB-based navigation. The legend has been omitted due to limited space, with the colors representing, from left to right in each group, anchors separated by 0.6\,m, 1.2\,m, 3\,m, 4\,m, 12\,m and 16\,m.}
    \label{fig:square_sim}
\end{figure*}   

\section{Experimental Results}

In this sections, we study the performance of the UWB-based cooperative localization system both in simulation and field experiments. 


\begin{figure}
    \centering
    \begin{subfigure}{0.49\textwidth}
        \centering
        \setlength\figureheight{0.42\textwidth}
        \setlength\figurewidth{0.95\textwidth}
        \footnotesize{\input{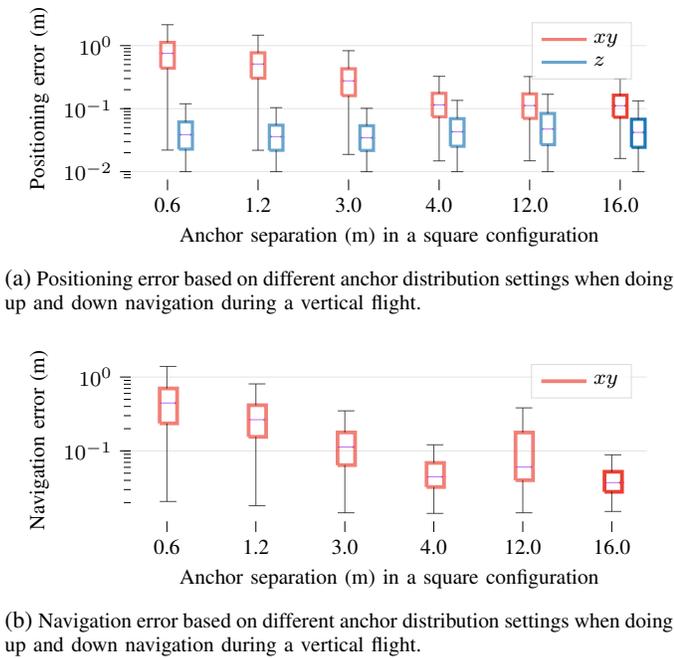}}
        \caption{Positioning error based on different anchor distribution settings when doing up and down navigation during a vertical flight.}
        \label{fig:up_down_pos}
    \end{subfigure}
    
    \vspace{1em}
    \begin{subfigure}{0.49\textwidth}
        \centering
        \setlength\figureheight{0.42\textwidth}
        \setlength\figurewidth{0.95\textwidth}
        \footnotesize{
\begin{tikzpicture}


\definecolor{color0}{rgb}{0.91796875,0.25,0.203125}
\definecolor{color1}{rgb}{0.6328125,0.203125,0.91796875}
\definecolor{color2}{rgb}{0.12156862745098,0.466666666666667,0.705882352941177}
\definecolor{color3}{rgb}{0.580392156862745,0.403921568627451,0.741176470588235}
\definecolor{color4}{rgb}{0.890196078431372,0.466666666666667,0.76078431372549}
\definecolor{color5}{rgb}{0.854901960784314,0.647058823529412,0.125490196078431}
\definecolor{color6}{rgb}{0.980392156862745,0.501960784313725,0.447058823529412}
\definecolor{color7}{rgb}{0.627450980392157,0.32156862745098,0.176470588235294}

\begin{axis}[
    height=\figureheight,
    width=\figurewidth,
    axis line style={white},
    tick align=outside,
    tick pos=left,
    x grid style={white!69.0196078431373!black},
    xtick style={color=black},
    y grid style={white!90!black},
    ymajorgrids,
    ytick style={color=black},
    scaled y ticks = false,
    legend cell align={left},
    legend style={
      fill opacity=0.8,
      draw opacity=0.666,
      text opacity=1,
      at={(0.97,0.97)},
      anchor=north east,
      draw=white!80!black
    },
    tick align=outside,
    tick pos=left,
    %
    %
    %
    %
    %
    %
    %
    log basis y={10},
    xlabel={Anchor separation (m) in a square configuration},
    xmin=3, xmax=32,
    xtick={5,10,15,20,25,30},
    xticklabels={0.6, 1.2, 3.0, 4.0, 12.0, 16.0},
    ylabel={Navigation error (m)},
    ymin=0.0113509597111723, ymax=1.74765612372479,
    ymode=log,
]
\addplot [line width=1.42pt, color0, forget plot, fill opacity=0.2, draw opacity=0.666]
table {%
4.5925 0.236205780787086
5.575 0.236205780787086
5.575 0.434148359697892
5.5375 0.44495072571635
5.575 0.455753091734807
5.575 0.70067693806158
4.5925 0.70067693806158
4.5925 0.455753091734807
4.59625 0.44495072571635
4.5925 0.434148359697892
4.5925 0.236205780787086
};
\addplot [black, forget plot, fill opacity=0.2, draw opacity=0.666]
table {%
5 0.236205780787086
5 0.0206895779141084
};
\addplot [black, forget plot, fill opacity=0.2, draw opacity=0.666]
table {%
5 0.70067693806158
5 1.3900411167484
};
\addplot [black, forget plot, fill opacity=0.2, draw opacity=0.666]
table {%
4.59625 0.0206895779141084
5.5375 0.0206895779141084
};
\addplot [black, forget plot, fill opacity=0.2, draw opacity=0.666]
table {%
4.59625 1.3900411167484
5.5375 1.3900411167484
};
\addplot [line width=1.42pt, color0, forget plot, fill opacity=0.2, draw opacity=0.666]
table {%
9.5925 0.155166271668099
10.575 0.155166271668099
10.575 0.258540842623703
10.5375 0.264733893231419
10.575 0.270926943839134
10.575 0.416108059253237
9.5925 0.416108059253237
9.5925 0.270926943839134
9.59625 0.264733893231419
9.5925 0.258540842623703
9.5925 0.155166271668099
};
\addplot [black, forget plot, fill opacity=0.2, draw opacity=0.666]
table {%
10 0.155166271668099
10 0.0181512185482159
};
\addplot [black, forget plot, fill opacity=0.2, draw opacity=0.666]
table {%
10 0.416108059253237
10 0.807418004645368
};
\addplot [black, forget plot, fill opacity=0.2, draw opacity=0.666]
table {%
9.59625 0.0181512185482159
10.5375 0.0181512185482159
};
\addplot [black, forget plot, fill opacity=0.2, draw opacity=0.666]
table {%
9.59625 0.807418004645368
10.5375 0.807418004645368
};
\addplot [line width=1.42pt, color0, forget plot, fill opacity=0.2, draw opacity=0.666]
table {%
14.5925 0.0642783958570472
15.575 0.0642783958570472
15.575 0.110207956216787
15.5375 0.113027699006243
15.575 0.1158474417957
15.575 0.178024406381381
14.5925 0.178024406381381
14.5925 0.1158474417957
14.59625 0.113027699006243
14.5925 0.110207956216787
14.5925 0.0642783958570472
};
\addplot [black, forget plot, fill opacity=0.2, draw opacity=0.666]
table {%
15 0.0642783958570472
15 0.0145732107322337
};
\addplot [black, forget plot, fill opacity=0.2, draw opacity=0.666]
table {%
15 0.178024406381381
15 0.348616418557662
};
\addplot [black, forget plot, fill opacity=0.2, draw opacity=0.666]
table {%
14.59625 0.0145732107322337
15.5375 0.0145732107322337
};
\addplot [black, forget plot, fill opacity=0.2, draw opacity=0.666]
table {%
14.59625 0.348616418557662
15.5375 0.348616418557662
};
\addplot [line width=1.42pt, color0, forget plot, fill opacity=0.2, draw opacity=0.666]
table {%
19.5925 0.0324889821559235
20.575 0.0324889821559235
20.575 0.0436106794015457
20.5375 0.0446527549668199
20.575 0.0456948305320942
20.575 0.0688315303029189
19.5925 0.0688315303029189
19.5925 0.0456948305320942
19.59625 0.0446527549668199
19.5925 0.0436106794015457
19.5925 0.0324889821559235
};
\addplot [black, forget plot, fill opacity=0.2, draw opacity=0.666]
table {%
20 0.0324889821559235
20 0.014271214002495
};
\addplot [black, forget plot, fill opacity=0.2, draw opacity=0.666]
table {%
20 0.0688315303029189
20 0.121083243093716
};
\addplot [black, forget plot, fill opacity=0.2, draw opacity=0.666]
table {%
19.59625 0.014271214002495
20.5375 0.014271214002495
};
\addplot [black, forget plot, fill opacity=0.2, draw opacity=0.666]
table {%
19.59625 0.121083243093716
20.5375 0.121083243093716
};
\addplot [line width=1.42pt, color0, forget plot, fill opacity=0.2, draw opacity=0.666]
table {%
24.5925 0.0401781042938873
25.575 0.0401781042938873
25.575 0.0571146579117238
25.5375 0.0606484390710363
25.575 0.0641822202303488
25.575 0.177422522725915
24.5925 0.177422522725915
24.5925 0.0641822202303488
24.59625 0.0606484390710363
24.5925 0.0571146579117238
24.5925 0.0401781042938873
};
\addplot [black, forget plot, fill opacity=0.2, draw opacity=0.666]
table {%
25 0.0401781042938873
25 0.0145664257391915
};
\addplot [black, forget plot, fill opacity=0.2, draw opacity=0.666]
table {%
25 0.177422522725915
25 0.382627442819525
};
\addplot [black, forget plot, fill opacity=0.2, draw opacity=0.666]
table {%
24.59625 0.0145664257391915
25.5375 0.0145664257391915
};
\addplot [black, forget plot, fill opacity=0.2, draw opacity=0.666]
table {%
24.59625 0.382627442819525
25.5375 0.382627442819525
};
\addplot [line width=1.42pt, color0]
table {%
29.5925 0.0279812838161758
30.575 0.0279812838161758
30.575 0.0364682033795934
30.5375 0.0372365183562849
30.575 0.0380048333329765
30.575 0.0522533752849953
29.5925 0.0522533752849953
29.5925 0.0380048333329765
29.59625 0.0372365183562849
29.5925 0.0364682033795934
29.5925 0.0279812838161758
};
\addlegendentry{$xy$}
\addplot [black, forget plot, fill opacity=0.2, draw opacity=0.666]
table {%
30 0.0279812838161758
30 0.0151678619195334
};
\addplot [black, forget plot, fill opacity=0.2, draw opacity=0.666]
table {%
30 0.0522533752849953
30 0.0885484219322979
};
\addplot [black, forget plot, fill opacity=0.2, draw opacity=0.666]
table {%
29.59625 0.0151678619195334
30.5375 0.0151678619195334
};
\addplot [black, forget plot, fill opacity=0.2, draw opacity=0.666]
table {%
29.59625 0.0885484219322979
30.5375 0.0885484219322979
};
\addplot [color1, forget plot, fill opacity=0.2, draw opacity=0.666]
table {%
4.59625 0.44495072571635
5.5375 0.44495072571635
};
\addplot [color1, forget plot, fill opacity=0.2, draw opacity=0.666]
table {%
9.59625 0.264733893231419
10.5375 0.264733893231419
};
\addplot [color1, forget plot, fill opacity=0.2, draw opacity=0.666]
table {%
14.59625 0.113027699006243
15.5375 0.113027699006243
};
\addplot [color1, forget plot, fill opacity=0.2, draw opacity=0.666]
table {%
19.59625 0.0446527549668199
20.5375 0.0446527549668199
};
\addplot [color1, forget plot, fill opacity=0.2, draw opacity=0.666]
table {%
24.59625 0.0606484390710363
25.5375 0.0606484390710363
};
\addplot [color1, forget plot, fill opacity=0.2, draw opacity=0.666]
table {%
29.59625 0.0372365183562849
30.5375 0.0372365183562849
};
\end{axis}

\end{tikzpicture}}
        \caption{Navigation error based on different anchor distribution settings when doing up and down navigation during a vertical flight.}
        \label{fig:up_down_nav}
    \end{subfigure}
    \caption{Positioning and navigation errors over a vertical flight to an altitude of 30\,m. The navigation error includes only the planar distance to the vertical line the drone is set to follow.}
    \label{fig:updown_sim}
\end{figure}

\subsection{Simulation Results}

The positioning and navigation errors for vertical flights are shown in Fig.~\ref{fig:updown_sim}. We observe that the positioning error consistently decreases as the anchors become more separated. For the small UGV setting, the error goes over 1\,m almost 20\% of the time, being highly unstable. It is worth noticing that the navigation error becomes relatively stable with the large UGV anchor distribution (1.2\,m separation). Navigation errors are in general lower than their positioning counterparts as the control of the drone is less affected by individual ranging errors, and these tend to average to zero as time passes. It is also worth noticing that the altitude error is significantly lower in all cases when compared to the planar $xy$ error.

Figure~\ref{fig:square_sim} then shows the results of flights following a square pattern. We can see that if UWB systems based on fixed anchors separated more than 10\,m are utilized, then the navigation error can be consistently maintained below 10\,cm. In the case of relying on small or large UGVs, the error is in the tens of centimeters, providing a competitive alternative to RTK-GNSS systems with higher deployment flexibility and lower system complexity.

\begin{figure}
    \centering
    \includegraphics[width=0.49\textwidth]{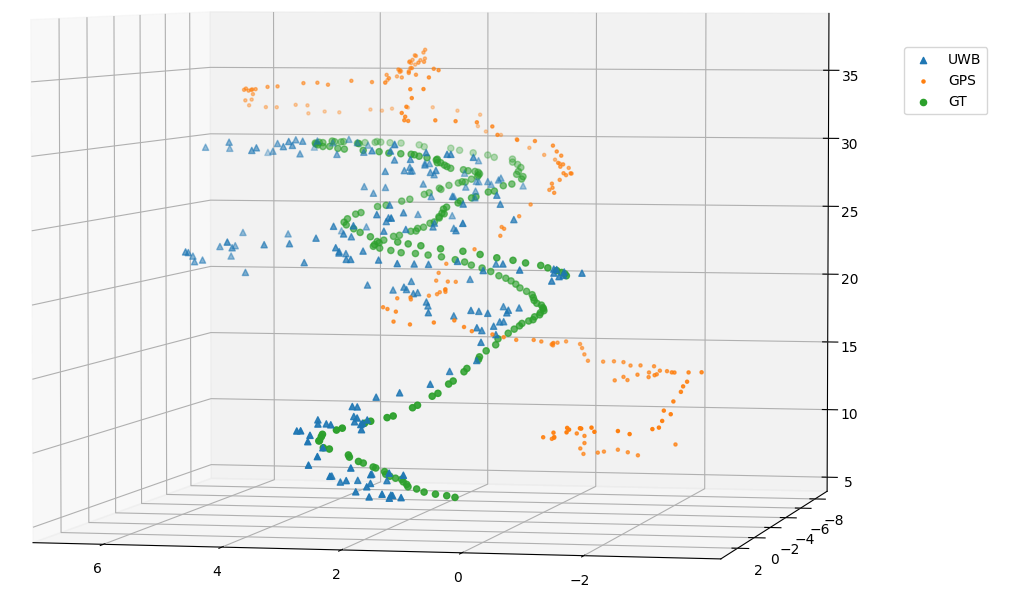}
    \caption{Partial trajectory of the UAV during the outdoors experiment. VIO is not included because it becomes unusable once the UAV reaches 8\,m of altitude.}
    \label{fig:trajectory}
\end{figure}

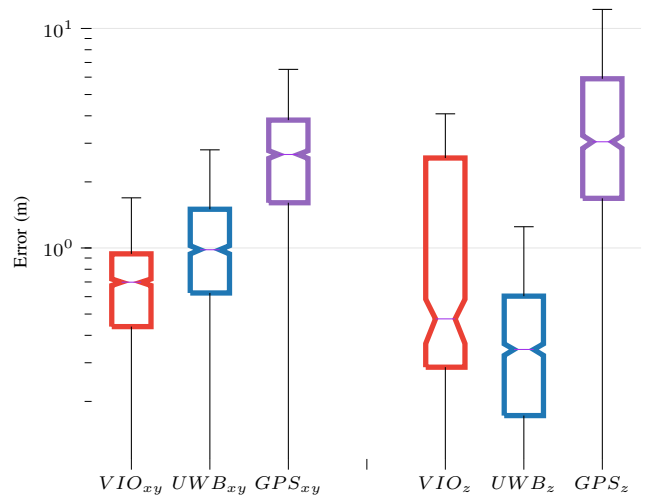
\begin{figure}
    \centering
    \setlength\figureheight{0.42\textwidth}
    \setlength\figurewidth{0.49\textwidth}
    \scriptsize{
\begin{tikzpicture}

\definecolor{color0}{rgb}{0.91796875,0.25,0.203125}
\definecolor{color1}{rgb}{0.6328125,0.203125,0.91796875}
\definecolor{color2}{rgb}{0.12156862745098,0.466666666666667,0.705882352941177}
\definecolor{color3}{rgb}{0.580392156862745,0.403921568627451,0.741176470588235}
\definecolor{color4}{rgb}{0.890196078431372,0.466666666666667,0.76078431372549}

\definecolor{color7}{rgb}{0.580392156862745,0.403921568627451,0.741176470588235}
\definecolor{color6}{rgb}{0.12156862745098,0.466666666666667,0.705882352941177}
\definecolor{color5}{rgb}{0.91796875,0.25,0.203125}

\begin{axis}[
    height=\figureheight,
    width=\figurewidth,
    axis line style={white},
    tick align=outside,
    tick pos=left,
    x grid style={white!69.0196078431373!black},
    xtick style={color=black},
    y grid style={white!90!black},
    ymajorgrids,
    ytick style={color=black},
    scaled y ticks = false,
    legend cell align={left},
    legend style={
      fill opacity=0.8,
      draw opacity=0.666,
      text opacity=1,
      at={(0.97,0.97)},
      anchor=north east,
      draw=white!80!black
    },
    tick align=outside,
    tick pos=left,
    %
    %
    %
    %
    %
    %
    %
    xmin=0.5, xmax=7.5,
    xtick={1,2,3,4,5,6,7},
    xticklabels={$VIO_{xy}$, $UWB_{xy}$, $GPS_{xy}$, , $VIO_z$, $UWB_z$, $GPS_z$},
    ylabel={Error (m)},
    ymajorgrids,
    ymin=0.110018668126381, ymax=12.822998030654,
    ymode=log,
]
\addplot [line width=2pt, color0, forget plot]
table {%
0.75 0.438256200984864
1.25 0.438256200984864
1.25 0.674593597279193
1.125 0.698461927361191
1.25 0.722330257443189
1.25 0.940177507410002
0.75 0.940177507410002
0.75 0.722330257443189
0.875 0.698461927361191
0.75 0.674593597279193
0.75 0.438256200984864
};
\addplot [black, forget plot]
table {%
1 0.438256200984864
1 0.0632751893602295
};
\addplot [black, forget plot]
table {%
1 0.940177507410002
1 1.69229048091101
};
\addplot [black, forget plot]
table {%
0.875 0.0632751893602295
1.125 0.0632751893602295
};
\addplot [black, forget plot]
table {%
0.875 1.69229048091101
1.125 1.69229048091101
};
\addplot [line width=2pt, color2, forget plot]
table {%
1.75 0.623723617389486
2.25 0.623723617389486
2.25 0.939113143679274
2.125 0.981610969574021
2.25 1.02410879546877
2.25 1.50084906283514
1.75 1.50084906283514
1.75 1.02410879546877
1.875 0.981610969574021
1.75 0.939113143679274
1.75 0.623723617389486
};
\addplot [black, forget plot]
table {%
2 0.623723617389486
2 0.0267978777144759
};
\addplot [black, forget plot]
table {%
2 1.50084906283514
2 2.79929851920673
};
\addplot [black, forget plot]
table {%
1.875 0.0267978777144759
2.125 0.0267978777144759
};
\addplot [black, forget plot]
table {%
1.875 2.79929851920673
2.125 2.79929851920673
};
\addplot [line width=2pt, color3, forget plot]
table {%
2.75 1.60703157784121
3.25 1.60703157784121
3.25 2.55844096266862
3.125 2.6658632409173
3.25 2.77328551916597
3.25 3.8241523758089
2.75 3.8241523758089
2.75 2.77328551916597
2.875 2.6658632409173
2.75 2.55844096266862
2.75 1.60703157784121
};
\addplot [black, forget plot]
table {%
3 1.60703157784121
3 0.0147874407403017
};
\addplot [black, forget plot]
table {%
3 3.8241523758089
3 6.51091049405551
};
\addplot [black, forget plot]
table {%
2.875 0.0147874407403017
3.125 0.0147874407403017
};
\addplot [black, forget plot]
table {%
2.875 6.51091049405551
3.125 6.51091049405551
};
\addplot [line width=2pt, color4, forget plot]
table {%
3.75 nan
4.25 nan
4.25 nan
4.125 nan
4.25 nan
4.25 nan
3.75 nan
3.75 nan
3.875 nan
3.75 nan
3.75 nan
};
\addplot [black, forget plot]
table {%
4 nan
4 nan
};
\addplot [black, forget plot]
table {%
4 nan
4 nan
};
\addplot [black, forget plot]
table {%
3.875 nan
4.125 nan
};
\addplot [black, forget plot]
table {%
3.875 nan
4.125 nan
};
\addplot [line width=2pt, color5, forget plot]
table {%
4.75 0.2863183199475
5.25 0.2863183199475
5.25 0.366798369904502
5.125 0.47536206777
5.25 0.583925765635498
5.25 2.5692779101575
4.75 2.5692779101575
4.75 0.583925765635498
4.875 0.47536206777
4.75 0.366798369904502
4.75 0.2863183199475
};
\addplot [black, forget plot]
table {%
5 0.2863183199475
5 0.0066105721099996
};
\addplot [black, forget plot]
table {%
5 2.5692779101575
5 4.08462203046
};
\addplot [black, forget plot]
table {%
4.875 0.0066105721099996
5.125 0.0066105721099996
};
\addplot [black, forget plot]
table {%
4.875 4.08462203046
5.125 4.08462203046
};
\addplot [line width=2pt, color6, forget plot]
table {%
5.75 0.17235825
6.25 0.17235825
6.25 0.324021683390278
6.125 0.344904499999998
6.25 0.365787316609718
6.25 0.603364999999999
5.75 0.603364999999999
5.75 0.365787316609718
5.875 0.344904499999998
5.75 0.324021683390278
5.75 0.17235825
};
\addplot [black, forget plot]
table {%
6 0.17235825
6 0.000573000000000157
};
\addplot [black, forget plot]
table {%
6 0.603364999999999
6 1.249623
};
\addplot [black, forget plot]
table {%
5.875 0.000573000000000157
6.125 0.000573000000000157
};
\addplot [black, forget plot]
table {%
5.875 1.249623
6.125 1.249623
};
\addplot [line width=2pt, color7, forget plot]
table {%
6.75 1.68318462421337
7.25 1.68318462421337
7.25 2.84204411227774
7.125 3.0463188178633
7.25 3.25059352344885
7.25 5.89927187944771
6.75 5.89927187944771
6.75 3.25059352344885
6.875 3.0463188178633
6.75 2.84204411227774
6.75 1.68318462421337
};
\addplot [black, forget plot]
table {%
7 1.68318462421337
7 0.0253823215293938
};
\addplot [black, forget plot]
table {%
7 5.89927187944771
7 12.2124063625276
};
\addplot [black, forget plot]
table {%
6.875 0.0253823215293938
7.125 0.0253823215293938
};
\addplot [black, forget plot]
table {%
6.875 12.2124063625276
7.125 12.2124063625276
};
\addplot [color1, forget plot]
table {%
0.875 0.698461927361191
1.125 0.698461927361191
};
\addplot [color1, forget plot]
table {%
1.875 0.981610969574021
2.125 0.981610969574021
};
\addplot [color1, forget plot]
table {%
2.875 2.6658632409173
3.125 2.6658632409173
};
\addplot [color1, forget plot]
table {%
3.875 nan
4.125 nan
};
\addplot [color1, forget plot]
table {%
4.875 0.47536206777
5.125 0.47536206777
};
\addplot [color1, forget plot]
table {%
5.875 0.344904499999998
6.125 0.344904499999998
};
\addplot [color1, forget plot]
table {%
6.875 3.0463188178633
7.125 3.0463188178633
};
\end{axis}

\end{tikzpicture}}
    \caption{Planar and vertical errors of the different methods during the outdoors flight. The VIO has low error but was only valid for the first few seconds of flight.}
    \label{fig:uwb_vio_gps_err}
\end{figure}

\subsection{Experimental results}

Results from outdoors experiments with real robots are reported in Fig.~\ref{fig:trajectory} and Fig.~\ref{fig:uwb_vio_gps_err}. The former shows a partial extract from the trajectory in 3D, where we can observe that the UWB error is significantly smaller even when the altitude reaches 30\,m. In the latter plot we can see that the overall error more than 5\,min flight time. The cooperative UWB approach particularly outperforms both VIO and GNSS estimations in terms of vertical accuracy. In terms of planar $xy$ error, VIO is more accurate but only during the first few seconds of flight, before it rapidly loses accuracy and diverges when the UAV altitude increases. In any case, the cooperative UWB-based localization provides consistent accuracy throughout the flight and therefore has potential for better collection of aerial data through autonomous flights.




\section{Conclusion}\label{sec:conclusion}

We have presented an analysis on how UWB-based relative localization between a UAV and a companion ground robot can improve the accuracy of autonomous flights outdoors. In particular, we have simulated different scenarios to assess the accuracy of the UWB-based relative positioning method. We have then validated this with robots in outdoor experiments, in an urban area where GNSS receivers do not perform optimally. Our analysis includes VIO estimation, which is more accurate at first but loses the reference when the UAV starts gaining altitude, presumably due to the lack of reference points.

In summary, we can conclude that UWB-based positioning systems can provide an alternative to RTK-GNSS when the accuracy of standalone GNSS is not enough for gathering aerial data. Moreover, we have proved that even when the transceivers are placed near each other in the ground, mounted on a mobile platform, the accuracy is enough to enable autonomous flight. 

Future work will focus on integrating UWB with GNSS data and performing experiments in more varied environments. We will also analyze different anchor configurations and consider additional robots.


\section*{Acknowledgment}

This research work is supported by the Academy of Finland's AutoSOS project (Grant No. 328755) and RoboMesh project (Grant No. 336061).

\bibliographystyle{unsrt}
\bibliography{bibliography}

\end{document}